\documentclass[11pt]{article}

\usepackage[margin=1in]{geometry}

\usepackage{graphicx}
\usepackage{booktabs}
\usepackage{array}
\usepackage{multirow}
\usepackage{xcolor}
\usepackage{tabularx}
\usepackage{placeins}

\usepackage{amsmath}
\usepackage{amssymb}
\usepackage{bm}

\usepackage{algorithm}
\usepackage{algpseudocode}
\usepackage{wrapfig}
\usepackage{tikz}
\usetikzlibrary{positioning,fit,calc,arrows.meta}

\definecolor{sourcecolor}{RGB}{185,110,70}
\definecolor{kvcolor}{RGB}{65,105,150}
\definecolor{windowcolor}{RGB}{110,120,135}
\definecolor{answercolor}{RGB}{45,120,75}

\usepackage{hyperref}

\usepackage[
    backend=biber,
    style=authoryear,
    maxcitenames=2,
    maxbibnames=99,
    labelnumber=true
]{biblatex}

\DeclareCiteCommand{\parencite}[\mkbibparens]
  {\usebibmacro{prenote}}
  {%
   \printnames{labelname}%
   \setunit{\nameyeardelim}%
   \printfield{year}%
   \addspace
   \mkbibbrackets{%
     \bibhyperref{\printfield{labelnumber}}%
   }%
  }
  {\multicitedelim}
  {\usebibmacro{postnote}}

\usepackage{tikz}
\usetikzlibrary{arrows.meta,calc,positioning}
\usetikzlibrary{fit}

\definecolor{embedblue}{RGB}{70,110,190}
\definecolor{lightdep}{RGB}{175,175,175}
\definecolor{infocolor}{RGB}{220,145,20}

\title{
    \textbf{Thinking Outside the Box:}\\
    Retention and Transmission of Information \\in Sliding-Window KV Inference
}

\author{
    Timothy DeLise\thanks{Corresponding author:
    \texttt{timothy.delise@umontreal.ca}}
    \qquad Seth Cromelin\\[0.5em]
}

\date{August 2026}

\begin{document}

\maketitle





\begin{abstract}
Sliding-window KV inference refers to processing a sequence
incrementally while retaining only a fixed-size cache of recent
key and value states. It can be applied to pretrained causal transformers at inference time without additional training, while its KV-cache memory remains fixed
as more tokens are processed. Because cached states are computed in
the context of earlier tokens, they may carry information from beyond
the current window and transmit it to later states.
This study presents a series of experiments using five open-weight
models spanning Qwen, Llama, Mistral, and Muse Glimmer. We investigate whether information originating outside the immediate context window can persist through a rolling KV cache and remain useful for retrieval. Initial results show that
retaining previously computed states improves retrieval across the
models tested compared with recomputing the final fixed window
from raw tokens. We then measure how far this effect extends and
find that Muse Glimmer and Mistral 7B show the strongest
\emph{latent information relay}: they can recover information
even after the relevant source tokens have left the cache. Both models incorporate sliding-window attention in their published
architectures, an association that motivates testing whether training
with sliding windows promotes more reliable information retention.
\end{abstract}

\section{Introduction}


This study investigates \emph{sliding-window KV inference}, autoregressive
inference using a fixed-width key-value (KV) cache. At each generation step,
the oldest cached position is discarded, a new position is appended, and
activations are computed only for the newly appended token. Variants of this
mechanism are used in models such as Mistral 7B
\parencite{jiang2023mistral}, which uses sliding-window attention throughout
the model, and Muse Glimmer \parencite{meta2026museglimmer}, which
interleaves sliding-window and full-attention layers. We evaluate information persistence under Rolling-KV inference
across five models: Qwen2.5 0.5B and 3B \parencite{qwen2025qwen25},
Mistral 7B \parencite{jiang2023mistral}, Llama 3.1 8B
\parencite{dubey2024llama}, and Muse Glimmer 30B
\parencite{meta2026museglimmer}, spanning multiple model architectures
and scales. Importantly, the
rolling KV cache is imposed at inference time on an otherwise
standard pretrained causal transformer, without requiring additional pretraining or architectural modification. The appeal of sliding-window KV
inference is that it combines a fixed memory footprint with the possibility
that representations inside the current window may carry information
originating from tokens that have already left the cache. In this sense, the
transformer becomes a bounded-state recurrent computation: past information
need not remain explicitly accessible if it can be propagated forward through
successive token representations. This raises the possibility of a simple
streaming approach to long-context inference. We test this hypothesis directly,
asking whether information originating outside the current KV window can
propagate into it, how much information survives, and over what distances. We
study these questions through three experiments comparing sliding-window KV
inference with standard fixed-window inference.

\begin{figure*}[]
\centering
\begin{tikzpicture}[
    x=1.05cm,
    y=1.15cm,
    >={Stealth[length=2mm]},
    state/.style={
        circle,
        draw=black!45,
        fill=black!3,
        minimum size=5mm,
        inner sep=0pt,
        line width=0.7pt
    },
    emb/.style={
        circle,
        draw=embedblue!85,
        fill=embedblue!10,
        minimum size=5mm,
        inner sep=0pt,
        line width=0.9pt
    },
    dep/.style={->, draw=lightdep, line width=0.5pt, opacity=0.35},
    highlightdep/.style={->, draw=black, line width=0.9pt},
    infoflow/.style={->, draw=infocolor, dashed, line width=0.9pt},
    vflow/.style={->, draw=embedblue!70, line width=0.8pt, opacity=0.8},
    outbox/.style={draw=black!70, rounded corners=2pt, minimum height=8mm, minimum width=18mm, align=center, inner sep=2pt},
    lab/.style={font=\small},
    smalllab/.style={font=\scriptsize},
    shiftstate/.style={
        circle,
        draw=shiftwindow!85,
        fill=shiftwindow!8,
        minimum size=5mm,
        inner sep=0pt,
        line width=0.8pt
    },
    shiftstatelight/.style={
        circle,
        draw=shiftwindow!85,
        fill=shiftwindow!8,
        minimum size=5mm,
        inner sep=0pt,
        line width=0.8pt,
        opacity=0.45
    },
    shiftdep/.style={
        ->,
        draw=shiftwindow!55,
        line width=0.55pt,
        opacity=0.55
    },
    legendlab/.style={
        font=\scriptsize,
        align=left
    },
]

\def\Tmax{7}      
\def\Lmax{4}      

\foreach \t in {0,...,7} {
    \node[emb]   (h\t0) at (\t,0) {};
    \node[state] (h\t1) at (\t,1) {};
    \node[state] (h\t2) at (\t,2) {};
    \node[state] (h\t3) at (\t,3) {};
    \node[state] (h\t4) at (\t,4) {};
}

%
\foreach \r/\rp in {0/1,1/2,2/3,3/4} {
    \foreach \t in {0,...,7} {
        \foreach \s in {0,...,\t} {
            \draw[dep] (h\s\r) -- (h\t\rp);
        }
    }
}


\node[text=infocolor, font=\Large] at (h52) {$\star$};
\node[text=infocolor, font=\Large] at (h00) {$\star$};

\foreach \s in {0,...,5} {
    \draw[highlightdep] (h\s1) -- (h52);
}


\draw[infoflow] (h00) .. controls +(0.7,0.5) and +(-1.0,-0.6) .. (h21);
\draw[infoflow] (h21) .. controls +(0.8,0.5) and +(-1.0,-0.5) .. (h52);

\node[lab, anchor=east] at (-0.95,0) {$x_i^{(1)}$: embeddings};
\node[lab, anchor=east] at (-0.95,1) {$x_i^{(2)}$};
\node[lab, anchor=east] at (-0.95,2) {$x_i^{(3)}$};
\node[lab, anchor=east] at (-0.95,3) {$\vdots$};
\node[lab, anchor=east] at (-0.95,4) {$x_i^{(L+1)}$};

\foreach \t [evaluate=\t as \ti using int(\t+1)] in {0,...,7} {
    \node[smalllab, anchor=north] at (\t,-0.350) {$t_{\ti}$};
}

\draw[->, line width=0.9pt] (-0.7,-0.75) -- (7.6,-0.75);
\node[lab, anchor=west] at (.8,-1) {token position / time $i$};

\draw[->, line width=0.9pt] (-0.75,0) -- (-0.75,4.3);
\node[lab, rotate=90] at (-2.1,2.0) {transformer depth $l$};

\draw[rounded corners=2pt, line width=1.0pt]
    (-0.5,-0.35) rectangle (7.5,4.55);

\node[font=\bfseries\small] at (3.6,4.9) {fixed context window $W = 8$};

\definecolor{shiftwindow}{RGB}{110,125,150}
\draw[shiftwindow!60, dashed, rounded corners=2pt, line width=0.8pt]
    (0.5,-0.35) rectangle (8.5,4.55);
\node[lab, text=shiftwindow!90, anchor=east] at (11.55,-1) {window advances one token};


\node[shiftstatelight] (g80) at (8,0) {};
\node[shiftstatelight] (g81) at (8,1) {};
\node[shiftstatelight] (g82) at (8,2) {};
\node[shiftstatelight] (g83) at (8,3) {};
\node[shiftstatelight] (g84) at (8,4) {};

\node[smalllab, text=shiftwindow!90, anchor=north] at (8,-0.350) {$t_{9}$};

\foreach \s in {0,...,7} {
    \draw[shiftdep] (h\s0) -- (g81);
}
\foreach \s in {0,...,7} {
    \draw[shiftdep] (h\s1) -- (g82);
}
\foreach \s in {0,...,7} {
    \draw[shiftdep] (h\s2) -- (g83);
}
\foreach \s in {0,...,7} {
    \draw[shiftdep] (h\s3) -- (g84);
}

\node[
    lab,
    anchor=west,
    fill=white,
    inner sep=1.5pt,
    rounded corners=1pt
] at (5.45,2.05)
{$x_{6}^{(l+1)} = F_{l}\!\left(x_{1:6}^{(l)}\right)$};

\node[
    outbox,
    anchor=south,
    minimum width=18mm
] (proj) at ($(h74.north)+(0,0.65)$)
{output\\projection};

\node[
    outbox,
    anchor=south,
    minimum width=18mm
] (soft) at ($(proj.north)+(0,0.35)$)
{softmax};

\node[
    lab,
    anchor=south
] (pred) at ($(soft.north)+(0,0.35)$)
{$p(t_{9}\mid t_{\leq 8})$};

\draw[->, line width=0.9pt] (h74.north) -- (proj.south);
\draw[->, line width=0.9pt] (proj.north) -- (soft.south);
\draw[->, line width=0.9pt] (soft.north) -- (pred.south);

\node[
    smalllab,
    anchor=south,
    fill=white,
    inner sep=1pt
] at (h74.east) {$x_{8}^{(L+1)}$};


\begin{scope}[xshift=-0.75cm, yshift=1cm]

\node[legendlab, text=black, anchor=west] (leg1) at (10.20,2.45)
    {causal dependence\\(highlighted example)};

\node[legendlab, text=black, anchor=west] (leg2) at (10.20,1.55)
    {causal dependence\\(all other states)};

\node[legendlab, text=black, anchor=west] (leg3) at (10.20,0.75)
    {example information flow\\(origin $\rightarrow$ later use)};

\node[legendlab, text=black, anchor=west] (leg4) at (10.20,-0.10)
    {information marker};

\node[text=infocolor, font=\large] (legstar) at (9.62,-0.10)
    {$\star$};

\node[
    fit=(leg1)(leg2)(leg3)(leg4)(legstar),
    draw=black!35,
    line width=0.35pt,
    rounded corners=2pt,
    fill=white,
    fill opacity=0.50,
    inner xsep=5mm,
    inner ysep=3mm
] (legendbox) {};

\draw[->, black, line width=0.7pt]
    (9.25,2.45) -- (10.00,2.45);

\draw[
    ->,
    draw=lightdep,
    line width=0.45pt,
    opacity=0.55
] (9.25,1.55) -- (10.00,1.55);

\draw[infoflow, line width=0.7pt]
    (9.25,0.75) -- (10.00,0.75);

\node[legendlab, text=black, anchor=west] (leg1_) at (10.20,2.45)
    {causal dependence\\(highlighted example)};

\node[legendlab, text=black, anchor=west] (leg2_) at (10.20,1.55)
    {causal dependence\\(all other states)};

\node[legendlab, text=black, anchor=west] (leg3_) at (10.20,0.75)
    {example information flow\\(origin $\rightarrow$ later use)};

\node[legendlab, text=black, anchor=west] (leg4_) at (10.20,-0.10)
    {information marker};

\node[text=infocolor, font=\large] (legstar_) at (9.62,-0.10)
    {$\star$};

\end{scope}

\node[font=\small\bfseries, text=infocolor, align=center] at (0.0,-1.55)
    {information\\originates here};

\node[font=\small\bfseries, text=infocolor, align=center] at (5.0,-1.55)
    {information\\encoded here};

\draw[->, draw=infocolor, line width=0.8pt] (0.0,-1.2) -- (0.0,-.8);
\draw[->, draw=infocolor, line width=0.8pt] (5.0,-1.2) -- (5.0,-0.8);

\draw[embedblue, line width=0.8pt]
    (-0.25,-2) .. controls +(0,-0.15) and +(0,-0.15) .. (3.6,-2)
    .. controls +(0,-0.15) and +(0,-0.15) .. (7.45,-2);
\node[font=\small, text=embedblue] at (3.6,-2.5) {context window of width $W=8$};

\end{tikzpicture}

\caption{
\textbf{Causal computation within a fixed transformer context window.}
Columns index token position / time $i$ and rows index transformer depth $l$.
The bottom row contains token embeddings $x_i^{(1)}=\operatorname{Embed}(t_i)$; higher rows contain successive residual-stream activations of the same dimensionality.
Each activation $x_i^{(l+1)}$ is computed from $x_i^{(l)}$ together with causal attention over $x_{1:i}^{(l)}$, as defined in Section~\ref{sec:standard-transformer}.
The final activation at the most recent position, $x_8^{(L+1)}$, determines the next-token distribution through the output projection and softmax.
The star marks information originating at an earlier position (here $t_1$) that is incorporated into a later residual-stream activation (here $x_6^{(l+1)}$) via the network's causal computations.
The faded outline indicates the same fixed-width window after advancing by one token. Notice that after the context window shifts, the origin of the information (first star) is no longer inside the new context window.
}
\label{fig:rolling-kv-grid}
\end{figure*}
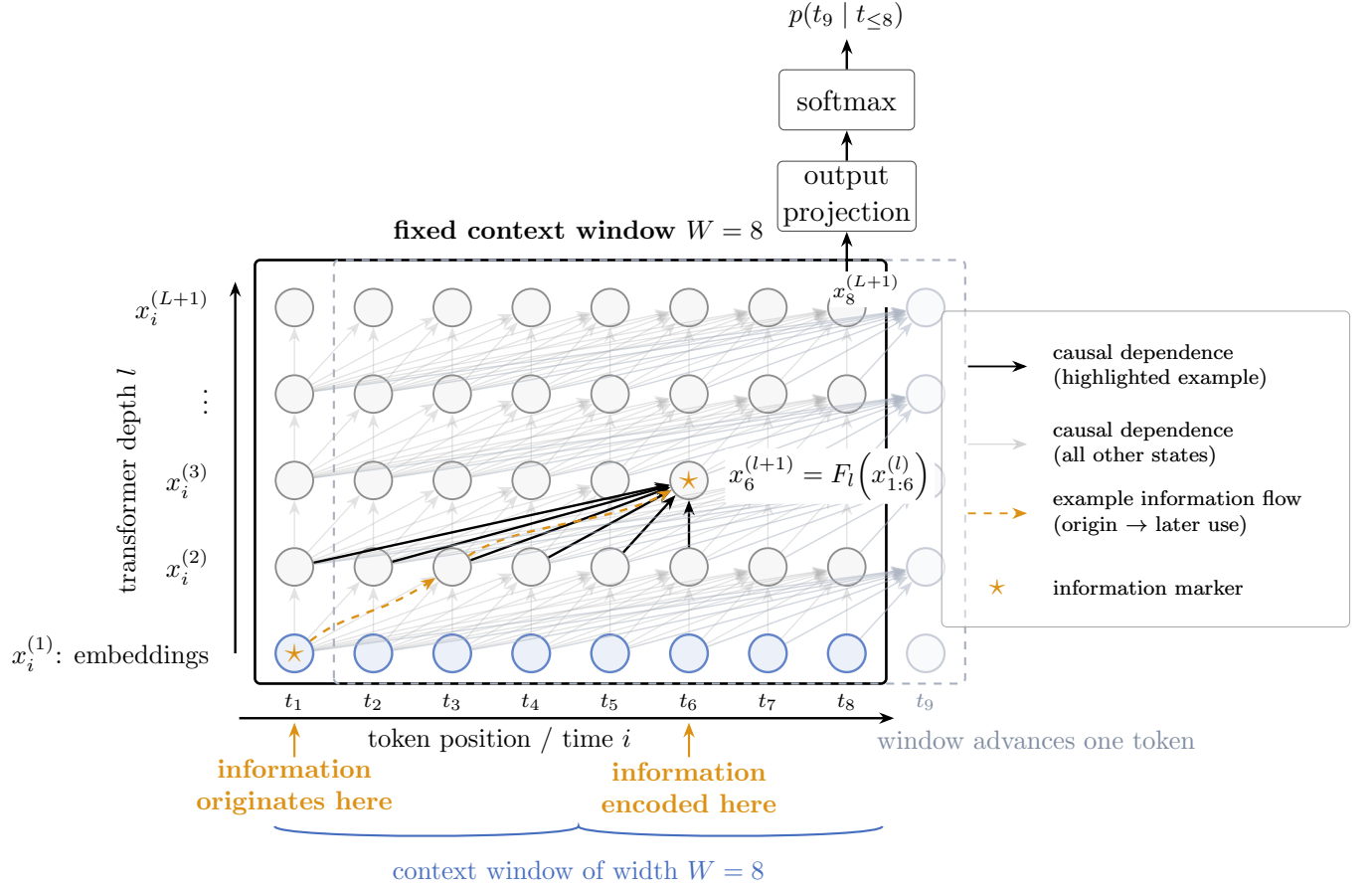

The mechanism underlying this possibility follows directly from causal
transformer computation \parencite{vaswani2017attention}. At each layer, attention is the only operation that
mixes information across token positions: the residual-stream activation at
position $i$ may incorporate information from positions $1,\ldots,i$, after
which the MLP acts independently at that position. Repeating this computation
through depth allows information originating at an earlier token to become
materialized in the residual-stream activations of later tokens.
Section~\ref{sec:standard-transformer} formalizes this computation, and
Figure~\ref{fig:rolling-kv-grid} illustrates the corresponding information
flow.

This creates a distinction between the position at which information
originates and the cached states in which its influence is subsequently
represented. Suppose information originating at token $t_i$ is incorporated
into the residual-stream activation of a later token $t_j$. 

Under ordinary fixed-window inference, once the window shifts and its contents are recomputed
without $t_i$, any dependence on that token is lost. Under sliding-window KV
inference, however, the already-computed cached state associated with $t_j$ is
retained. Thus, when $t_i$ is eventually evicted, information originating
there may remain materialized in cached states that are still inside the
window.

The central empirical question is therefore: how much of this information
survives, and for how long?

This question is relevant to the broader problem of long-context inference.
Standard transformer inference requires memory that grows with the amount of
retained context, whereas recurrent architectures maintain a state whose size
is independent of sequence length. State-space models such as Mamba
\parencite{gu2023mamba} have renewed interest in this recurrent paradigm,
while recent hybrid architectures retain attention alongside recurrent
state-space layers to balance bounded-memory sequence processing with the
capabilities of attention \parencite{nvidia2025nemotronh}. Sliding-window KV inference occupies an interesting
intermediate regime: its explicit state remains bounded, yet that state may
preserve information originating far in the past through successive propagation.
If such propagation is substantial, a pretrained transformer may exhibit useful 
recurrent memory even when only a fixed-width KV cache is retained.


\section{Related Work}

\paragraph{Information Flow and Mechanistic Interpretability}
Mechanistic interpretability work has explicitly attempted to reverse engineer
and deconstruct transformers in order to trace information flow and better
understand the inner workings of their components. Elhage et al.
\parencite{elhage2021mathematical} give a framework in which the residual
stream acts as a shared communication channel that components read from and
write to. Causal attention uses the residual stream to select or read
information from earlier or current token positions and writes a transformed
contribution to the residual stream of a destination token. Because attention
can write information from earlier token representations into the residual
stream, the current token representation can inherently encode more than just
the raw identity of the token. Meng et al.
\parencite{meng2022locating} give evidence that factual information can be
localized to middle-layer MLPs, and that targeted modification of these MLP
mappings can alter downstream factual predictions.

\paragraph{KV Caches and Long-context Inference}
KV caches preserve earlier computed key/value states in order to avoid later
recomputation; long-context methods then optimize which cached states to keep,
omit, or compress when memory is bounded. StreamingLLM
\parencite{xiao2023streamingllm} is one example showing that naive sliding
eviction of tokens can destabilize inference, and that retaining specific
initial KV states helps. Results from the softmax attention distribution show
that the initial tokens receive disproportionately high attention weights, a
phenomenon the authors term \emph{attention sinks}. StreamingLLM exploits this
behavior by retaining the KV states of several initial sink tokens alongside
the most recent tokens in the sliding window, which stabilizes attention
computation and prevents the large increase in perplexity caused by evicting
those initial states. Like StreamingLLM, H$_2$O
\parencite{zhang2023h2o} takes advantage of the fact that some cached tokens
matter more than others. The key observation is that a small subset of tokens
receive a disproportionate amount of cumulative attention. The authors call
these high-attention tokens \emph{heavy hitters}.

\paragraph{Recurrent and Compressed Transformer Memory}
Recurrent memory approaches extend the context that is usable by carrying
internal states from previous caches forward rather than recomputing them from
scratch. Transformer-XL \parencite{dai2019transformerxl} caches previous hidden
states and allows them to be reused in subsequent segments, enabling
dependencies beyond a fixed-length context. This differs from a rolling KV
setup where historical information may remain embedded in retained states
after older source tokens disappear. Rae et al.
\parencite{rae2019compressive} extend recurrent memory by compressing older
hidden states into a lower-resolution memory rather than discarding them,
allowing useful historical information to remain accessible over longer
ranges.

More recently, state-space models such as Mamba
\parencite{gu2023mamba} have renewed interest in recurrent sequence models
whose inference state does not grow with sequence length. At the same time,
several recent production-oriented architectures have adopted hybrid designs
rather than replacing attention entirely. NVIDIA's Nemotron-H
\parencite{nvidia2025nemotronh} replaces the majority of self-attention layers
with Mamba layers while retaining periodic full-attention layers, combining
constant-memory recurrent computation with direct attention over the context.
Muse Glimmer \parencite{meta2026museglimmer} follows a related design principle
within an attention-based architecture, repeating three 2048-token
sliding-window attention layers followed by one full-attention layer. These
hybrid architectures illustrate practical interest in obtaining as much of the
memory and inference benefit of bounded-state computation as possible while
retaining periodic global access to the context.

\section{Background and Inference Setup}

\subsection{Transformer Computation}
\label{sec:standard-transformer}

Let $x_i^{(l)} \in \mathbb{R}^{d_{\mathrm{model}}}$ denote the residual-stream
activation at token position $i$ at the start of layer $l$, with
$x_i^{(1)} = \operatorname{Embed}(t_i)$. Each layer applies causal attention
over the current and preceding positions, followed by an MLP:
\begin{align}
\tilde{x}_i^{(l)}
&=
x_i^{(l)}
+
\operatorname{Attn}^{(l)}
\left(x_{1:i}^{(l)}\right),
\\
x_i^{(l+1)}
&=
\tilde{x}_i^{(l)}
+
\operatorname{MLP}^{(l)}
\left(\tilde{x}_i^{(l)}\right).
\end{align}

The final state at position $i$ is mapped to vocabulary logits and a next-token
distribution:
\begin{equation}
p(t_{i+1}\mid t_{\leq i})
=
\operatorname{softmax}
\left(
\operatorname{Unembed}
\left(x_i^{(L+1)}\right)
\right).
\end{equation}

Figure~\ref{fig:rolling-kv-grid} visualizes the same computation as a grid over
token position and transformer depth, with the embedding states $x_i^{(1)}$ at
the bottom and causal information flow proceeding upward and to the right through successive residual-stream activations. 

Notation adapted from \parencite{arditi2024refusallanguagemodelsmediated}.

\subsection{Information Flow in Causal Attention}
\label{sec:information-propagation}

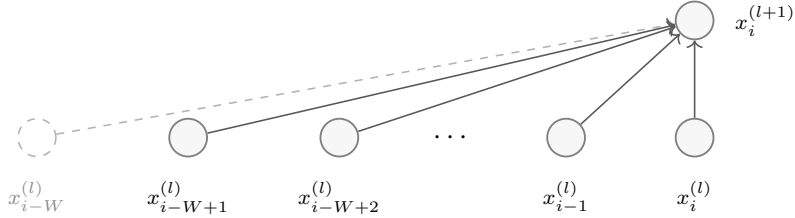
\begin{figure}[htbp]
    \centering
    \begin{tikzpicture}[
    x=1cm,
    y=1cm,
    state/.style={
        circle,
        draw=black!50,
        fill=black!3,
        minimum size=5mm,
        inner sep=0pt,
        line width=0.7pt
    },
    outside/.style={
        circle,
        draw=black!30,
        dashed,
        minimum size=5mm,
        inner sep=0pt,
        line width=0.6pt
    },
    dep/.style={
        ->,
        draw=black!65,
        line width=0.6pt
    },
    depdash/.style={
        ->,
        draw=black!30,
        dashed,
        line width=0.6pt
    }
]

\node[outside] (outside) at (0,0) {};

\node[state] (x1)   at (2.0,0) {};
\node[state] (x2)   at (4.0,0) {};
\node              at (5.5,0) {$\cdots$};
\node[state] (xim1) at (7.0,0) {};
\node[state] (xi)   at (8.7,0) {};

\node[state] (target) at (8.7,1.55) {};

\draw[depdash] (outside) -- (target);
\draw[dep] (x1) -- (target);
\draw[dep] (x2) -- (target);
\draw[dep] (xim1) -- (target);
\draw[dep] (xi) -- (target);

\node[below=5pt of outside, font=\scriptsize, text=black!45]
    {$x_{i-W}^{(l)}$};

\node[below=5pt of x1, font=\scriptsize]
    {$x_{i-W+1}^{(l)}$};

\node[below=5pt of x2, font=\scriptsize]
    {$x_{i-W+2}^{(l)}$};

\node[below=5pt of xim1, font=\scriptsize]
    {$x_{i-1}^{(l)}$};

\node[below=5pt of xi, font=\scriptsize]
    {$x_i^{(l)}$};

\node[right=4pt of target, font=\scriptsize]
    {$x_i^{(l+1)}$};

\end{tikzpicture}

    \caption{
    Single-layer information flow under causal attention with window width
    $W$. The state $x_i^{(l+1)}$ may depend on the $W$ states
    $x_{i-W+1:i}^{(l)}$. The dashed state lies outside the attention window
    and therefore has no direct path to the target at this layer.
    }
    \label{fig:single-layer-information-flow}
\end{figure}

The causal attention computation provides a simple way to reason about how
information can be transmitted forward in token position. At every layer,
there is a possible information transmission from past residual-stream
activations to a later residual-stream activation. Figure~
\ref{fig:single-layer-information-flow} isolates this single-layer dependency
from the larger computation shown in Figure~\ref{fig:rolling-kv-grid}.

For a fixed attention window of width $W$, the state at position $i$ and depth $l+1$ can directly attend to the $W$ positions

\begin{equation}
    i-W+1,\ldots,i
\end{equation}

of depth $l$.

Going from one layer to the next, information can be transmitted across
at most $W-1$ token-position steps. Using the notation of
Section~\ref{sec:standard-transformer}, the local-attention computation may be
written schematically as

\begin{equation}
x_i^{(l+1)}
=
F_l\left(x_{i-W+1:i}^{(l)}\right),
\label{eq:local-layer-map}
\end{equation}

where $F_l$ represents the complete computation of layer $l$, including the
attention operation, residual connections, and position-wise MLP.

The important point is that the inputs to $F_l$ are not raw tokens except at
the first layer. They are themselves residual-stream activations produced by
the preceding layers. Each of these states may therefore already contain
information originating at still earlier token positions. Information can only
be transmitted across token positions through attention, but it can be
transmitted again from layer $l$ to layer $l+1$. This layer-by-layer growth of the receptive field is also described
for Mistral 7B's sliding-window attention architecture
\parencite{jiang2023mistral}.

This gives a simple recursive description of the maximum possible information
flow. Let $R_l$ denote the greatest distance into the past from which
$x_i^{(l+1)}$ can have a causal dependence after $l$ attention layers. Before
any attention layer is applied, the embedding at position $i$ depends only on
token $t_i$, so

\begin{equation}
R_0 = 0.
\end{equation}

Each additional attention layer can extend the maximum possible
receptive field by $W-1$ token positions. Hence, for the dependency
graph,

\begin{equation}
R_{l+1}=R_l+(W-1),
\end{equation}

\begin{equation}
\boxed{R_L=L(W-1).}
\label{eq:receptive-field-bound}
\end{equation}

Equivalently, the final residual-stream activation at position $i$ can in
principle depend on input tokens as far back as

\begin{equation}
t_{\,i-L(W-1)}.
\end{equation}

The distinction between $W$ and $W-1$ here is only a matter of counting
positions versus distance: an attention window containing $W$ token positions
spans $W-1$ steps from its oldest position to its newest. Thus the receptive
field grows approximately as $LW$, but its exact maximum backward distance
under this convention is $L(W-1)$.

This is a statement about \emph{possible causal dependence}, not a guarantee
that useful information is actually transmitted. The architecture provides a
path by which information can move forward through successive residual-stream
activations; whether a pretrained transformer uses that path to preserve
behaviorally useful information, and how much information survives as the
source becomes more distant, are empirical questions.

This observation is especially relevant to sliding-window KV inference. When
the window advances, previously computed states are retained rather than
recomputed. A retained state may therefore carry information that originated
at an earlier position which is no longer directly available in the current
window. As new states are subsequently computed, that information may in
principle be transmitted forward again. The experiments in
Section~\ref{sec:experiments} test whether this potential information flow is
measurable in practice.

\subsection{Rolling-KV Inference}

We define Rolling-KV inference as a fixed width KV-Cache procedure where tokens are ingested sequentially 
such that the cache retains at most $W$ token positions at any given time. 
Once the cache reaches its fixed capacity, it continues to ingest new tokens sequentially while evicting the 
oldest cache position. For example, if $W = 4$, after the cache contains four positions and a fifth token arrives, the first token is evicted while the 2nd through 5th remain. But the key here is that the first token is only evicted once the fifth token has been processed using the entire cache. 
Importantly, when the window shifts the K/V cache is retained rather than recomputed from the tokens. This is because the retained K/V states were originally computed while earlier context was still available. 
Therefore, the surviving cache may contain residual information influenced by earlier processed tokens that no longer explicitly remain in the cache. 
%

\begin{algorithm}[h]
\caption{Rolling-KV Inference}
\begin{algorithmic}[1]

\Require model $M$, token sequence $t_{1:N}$, window width $W$
\State Process the first $W$ tokens with model $M$ and store the resulting K/V states as the initial cache.
\For{each remaining token position $W+1$ through $N$}
\State Process the current token using the retained K/V cache.
\State Add the current token's new K/V state to the cache.
\State If the cache now exceeds $W$ positions, then evict the oldest cache position. 
\EndFor
\Ensure final K/V cache containing at most $W$ token positions. 

\end{algorithmic}
\end{algorithm}

\section{Experiments}
\label{sec:experiments}

\subsection{Experimental Setup}

All three experiments evaluate the same five models:
Qwen2.5-0.5B-Instruct, Qwen2.5-3B-Instruct,
Mistral-7B-Instruct-v0.1, Meta-Llama-3.1-8B-Instruct, and
Muse-Glimmer-30B. The first four run in BF16; Glimmer uses NF4
quantization with BF16 computation. All experiments were run on a single NVIDIA RTX 3090 GPU
with 24 GB of memory. We use raw completion prompts
without chat templates, neutral filler, and a retained KV window of
$W=512$ token positions. No model is trained for these experiments.

We compare three inference conditions. 

\begin{enumerate}
    \item \textbf{Rolling-KV} processes the full
prompt in order, retaining at most the newest $W$ KV positions after
each step. Previously computed KV states are retained rather than
recomputed when the window advances.
    \item \textbf{Last Window }processes only the
final $W$ raw tokens, recomputing their states from scratch. 
    \item \textbf{Full Prompt} processes the entire prompt, subject to each model's native attention rules.
\end{enumerate}

For Glimmer, Rolling-KV crops the caches of both its local and global attention layers to $W$ positions.

Each trial has four model-specific, single-token candidate answers.
We score their next-token logits directly and count the trial as
correct when the intended answer receives the highest score; chance
accuracy is $25\%$. Candidates are not printed in the prompt.
Experiments 1 and 2 use 100 trials per offset, model, and condition,
with seed 1. Experiment 3 uses the trial counts described below. 

Code, experiment configurations, raw results, and figures are available
at \url{https://github.com/jefferythewind/Thinking-Outside-the-Box}.

\subsection{Experiment 1: Secret-Code Retrieval}

\paragraph{Task and Motivation}
The first experiment tests information persistence near the window
boundary. Each trial introduces a secret code with the phrase
``The secret code is:'', followed by filler that moves the code
toward and then beyond the final raw-window boundary. At the end,
the model is asked for the code, which it selects from four
candidate answer tokens.

The offset is the start index of the final $W$-token raw window
minus the code's token index. At offset $0$, the code is the first
token of that raw window. At $+1$, it has left the final raw window
but is still directly accessible during the last Rolling-KV
scoring step. At $+2$ and beyond, its own KV state is no longer
directly accessible at scoring time.

\begin{table}[htbp]
\centering
\small
\begin{tabularx}{\linewidth}{@{}X@{}}
\toprule
\textbf{Illustrative prompt structure} \\[0.25em]
{\ttfamily\raggedright
The secret code is: \fbox{ER}\quad
The document contains ordinary background notes about trees, roads,
weather, tools, books, chairs, \ldots\quad
Question: What is the secret code?\quad Answer:
\par}
\\[0.4em]
{\footnotesize
The boxed code is the anchor for the offset. The filler length
varies to place it at different positions relative to the final
raw window. Candidate answers are scored from next-token logits
and are not printed in the prompt.}
\\
\bottomrule
\end{tabularx}
\caption{Structure of a secret-code trial. The code and filler
vary between trials.}
\label{tab:secret-boundary}
\end{table}

\paragraph{Results}
We sweep offsets from $-5$ to $+5$. At offset $+1$, Rolling-KV
achieves $100\%$ accuracy for each of the five models, while the
equal-weight mean of the Last Window accuracies is $24.4\%$.
This contrast occurs while the secret code's KV state remains directly
accessible to the last Rolling-KV scoring step, although the
code is absent from the Last Window input.

At offset $+2$, after direct access to the code's KV state has
ended, the models diverge. Rolling-KV accuracy is $32\%$ for
Qwen 0.5B, $47\%$ for Qwen 3B, $82\%$ for Llama 8B, $97\%$
for Mistral 7B, and $100\%$ for Glimmer 30B. Mistral and
Glimmer remain at or above $95\%$ throughout the measured
offsets $+2$ through $+5$, whereas the Qwen models fall
substantially and Llama's advantage diminishes with distance.
Full Prompt remains near perfect across the sweep.

\begin{figure}[htbp]
    \centering
    \includegraphics[width=\linewidth]
        {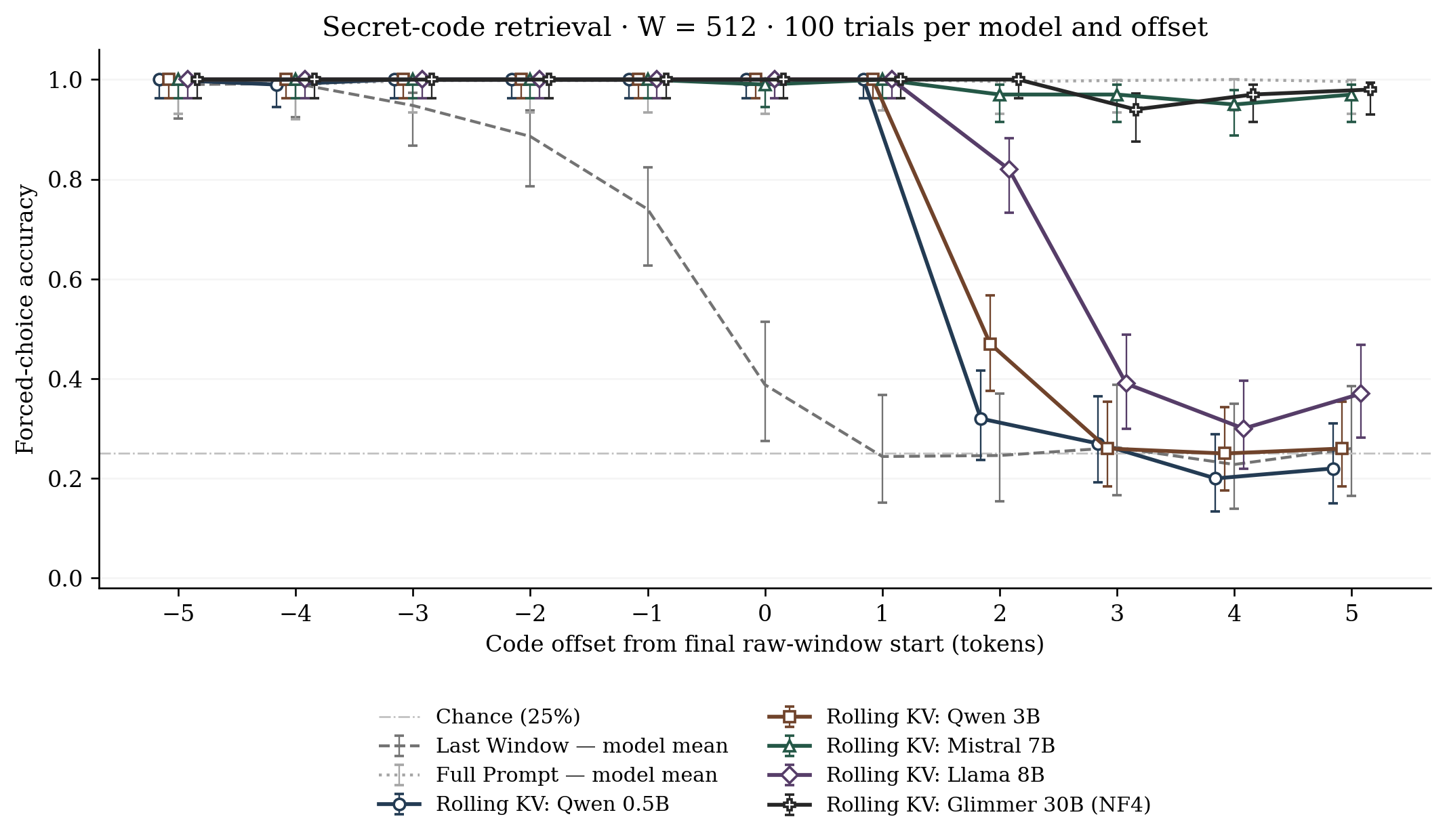}
    \caption{Secret-code retrieval accuracy across offsets $-5$
    through $+5$, with $W=512$. Each condition has 100 trials
    per offset and model. The combined control curves are
    equal-weight means across the five models; Rolling-KV curves
    are shown separately. At $+1$, the code is outside the final
    raw window but remains directly accessible during the last
    Rolling-KV scoring step. From $+2$ onward, its own KV state
    is no longer directly accessible. Four-choice chance
    accuracy is $25\%$.}
    \label{fig:secret-code-boundary}
\end{figure}

\FloatBarrier

\subsection{Experiment 2: Extended Secret-Code Retrieval}

\paragraph{Task and Motivation}
Experiment 2 uses the same boundary intervention but places the
code after a longer defining phrase: ``The secret code is defined
by the following identifier:''. This moves portions of the
definition out of the final raw window before the code reaches
its boundary. We again ask the model to select the code from four
candidate answer tokens.

\begin{table}[htbp]
\centering
\small
\begin{tabularx}{0.95\linewidth}{@{}lX@{}}
\toprule
\textbf{Task variant} & \textbf{Prompt prefix} \\
\midrule
Short code &
{\ttfamily The secret code is: \fbox{ER}\quad
The document contains ordinary background notes \ldots}
\\[0.5em]
Long definition &
{\ttfamily The secret code is defined by the following identifier:
\fbox{NG}\quad The document contains ordinary background notes
\ldots}
\\
\bottomrule
\end{tabularx}
\caption{Secret-code definitions used in Experiments 1 and 2.
Experiment 2 retains the same code-relative offset convention
while lengthening the defining phrase.}
\label{tab:secret-code-variants}
\end{table}

\paragraph{Results}
We sweep offsets from $-10$ to $+10$. As the defining phrase
leaves the final raw window, Last Window accuracy declines.
At offset $0$, its equal-weight mean across models is $40.2\%$,
compared with $95.8\%$ for Rolling-KV. At $+1$, the corresponding
accuracies are $27.0\%$ and $92.0\%$. Full Prompt remains near
perfect across the sweep.

Once the code's own KV state is no longer directly accessible,
retrieval again differs by model. At $+2$, Rolling-KV achieves
$30\%$ for Qwen 0.5B, $35\%$ for Qwen 3B, $79\%$ for Llama 8B,
$95\%$ for Mistral 7B, and $98\%$ for Glimmer 30B. At $+10$,
Mistral and Glimmer still achieve $89\%$ and $96\%$,
respectively. Llama achieves $56\%$ at that offset, while both
Qwen models are near chance. Thus, the longer definition
strengthens the distinction between the control and Rolling-KV
before eviction, while retrieval after the code's eviction
varies substantially across models.

\begin{figure}[htbp]
    \centering
    \includegraphics[width=\linewidth]
        {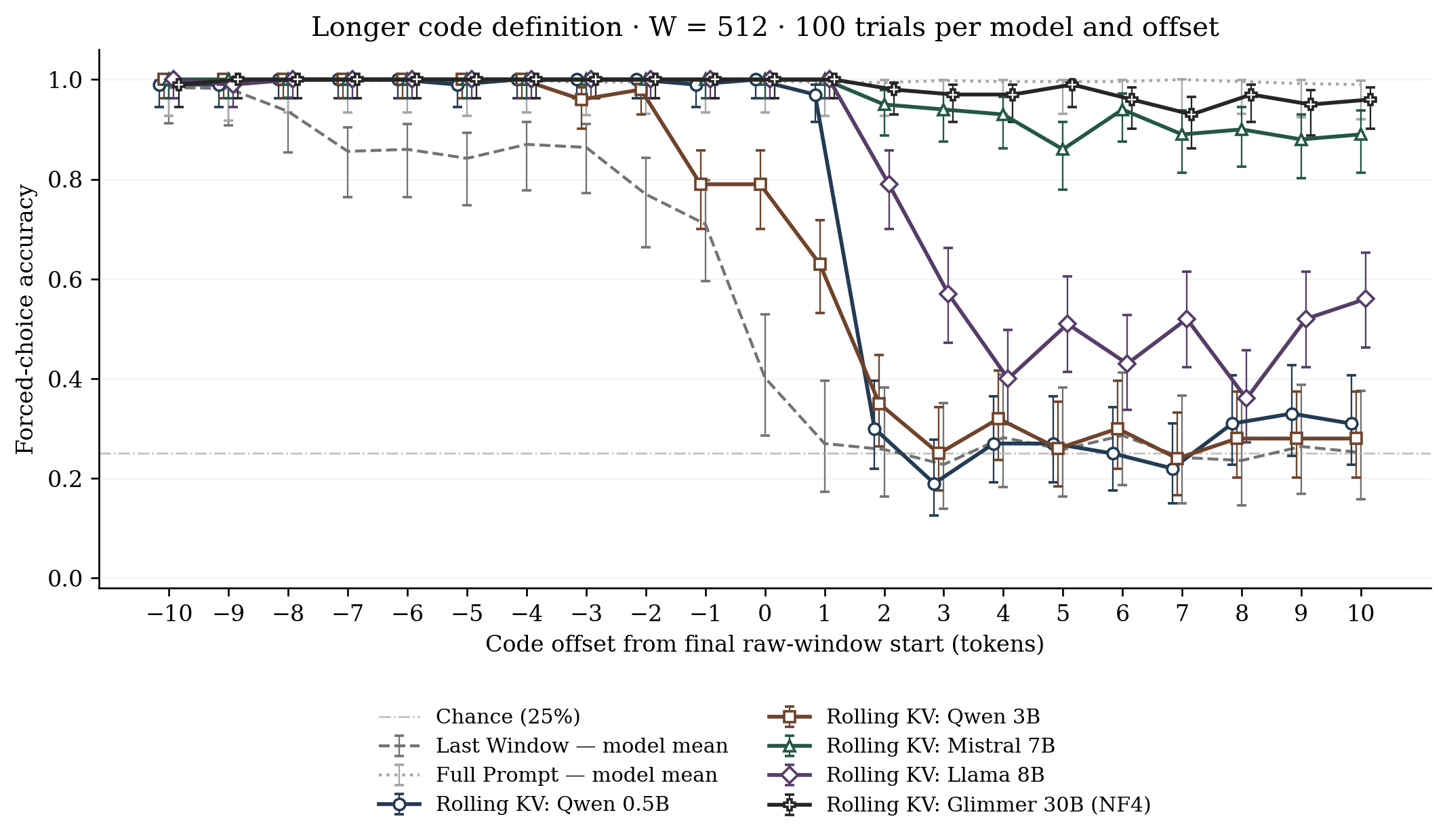}
    \caption{Forced-choice accuracy for the extended secret-code
    task across offsets $-10$ through $+10$, with $W=512$.
    Each condition has 100 trials per offset and model. Combined
    control curves are equal-weight means across models; Rolling-KV
    curves are shown separately. At $+1$, the code remains directly
    accessible to the final Rolling-KV scoring step. At $+2$ and
    beyond, its own KV state is no longer directly accessible.
    Four-choice chance accuracy is $25\%$.}
    \label{fig:extended-secret-code-accuracy}
\end{figure}

\FloatBarrier

\subsection{Experiment 3: Definition-to-Carrier Distance}
\label{sec:definition-to-carrier}

\paragraph{Task and Motivation}
The first two experiments use instructions of varying length to
define a secret code, then test retrieval as the definition and
code move across the final context-window boundary. Experiment 3
separates the definition from the code's later occurrence. An
instruction identifies an arbitrary marker indicating where the
secret will appear. After a controlled gap, the marker appears
again immediately before the code. We call this later
marker--code occurrence the \emph{carrier}.

This construction tests two distances independently: the gap
from the definition to the carrier, and the distance from the
code to the final window boundary. In particular, it lets us
ask whether the earlier definition remains useful when it is
outside the window available as the carrier is processed, and
whether the answer remains recoverable after the code itself
has also been evicted.

\begin{table}[htbp]
\centering
\small
\begin{tabularx}{0.92\linewidth}{@{}X@{}}
\toprule
\textbf{Illustrative prompt structure} \\[0.3em]
\begin{minipage}{\linewidth}
\ttfamily\raggedright
This is a memory test.\newline
The secret code will appear later in this text.\newline
The secret is the value immediately following
\textless{}Q3\textgreater{}.\newline
\ldots\ [definition-to-carrier filler] \ldots\newline
\textless{}Q3\textgreater{}:\fbox{NULL}\newline
\ldots\ [tail filler] \ldots\newline
Question: What is the secret code?\newline
Answer:
\par
\end{minipage}
\\[0.4em]
{\footnotesize
The definition-to-carrier gap counts filler tokens after the
defining preamble and before the later marker prefix. Tail filler
sets the boxed code's position relative to the final raw window.
Candidate answers are scored from next-token logits; they are
not listed in the prompt.}
\\
\bottomrule
\end{tabularx}
\caption{Structure of an Experiment 3 trial, illustrated with a
marker and answer found in the saved context examples. The marker
and code vary between trials.}
\label{tab:definition-carrier-example}
\end{table}

\FloatBarrier

\paragraph{Experimental Construction}
Each prompt begins with 128 filler tokens followed by the defining
preamble. We vary the number of filler tokens between the end of
that preamble and the carrier prefix over
$D\in\{32,64,128,256,512,768,1024\}$. The tokenized marker
prefix adds a few more tokens between the preamble and the code.

Independently, we vary the code-relative boundary offset over
$\{-32,0,32,64,128,256\}$. The offset is the start index of the
final $W$-token raw window minus the code's index. Thus, at $-32$
the code is 32 tokens inside that window; at $0$ it is the
window's first token. At the positive offsets in this sweep,
the code is absent from both the final raw window and the
retained Rolling-KV positions at scoring time.

\begin{figure}[htbp]
    \centering
    \includegraphics[width=\linewidth]{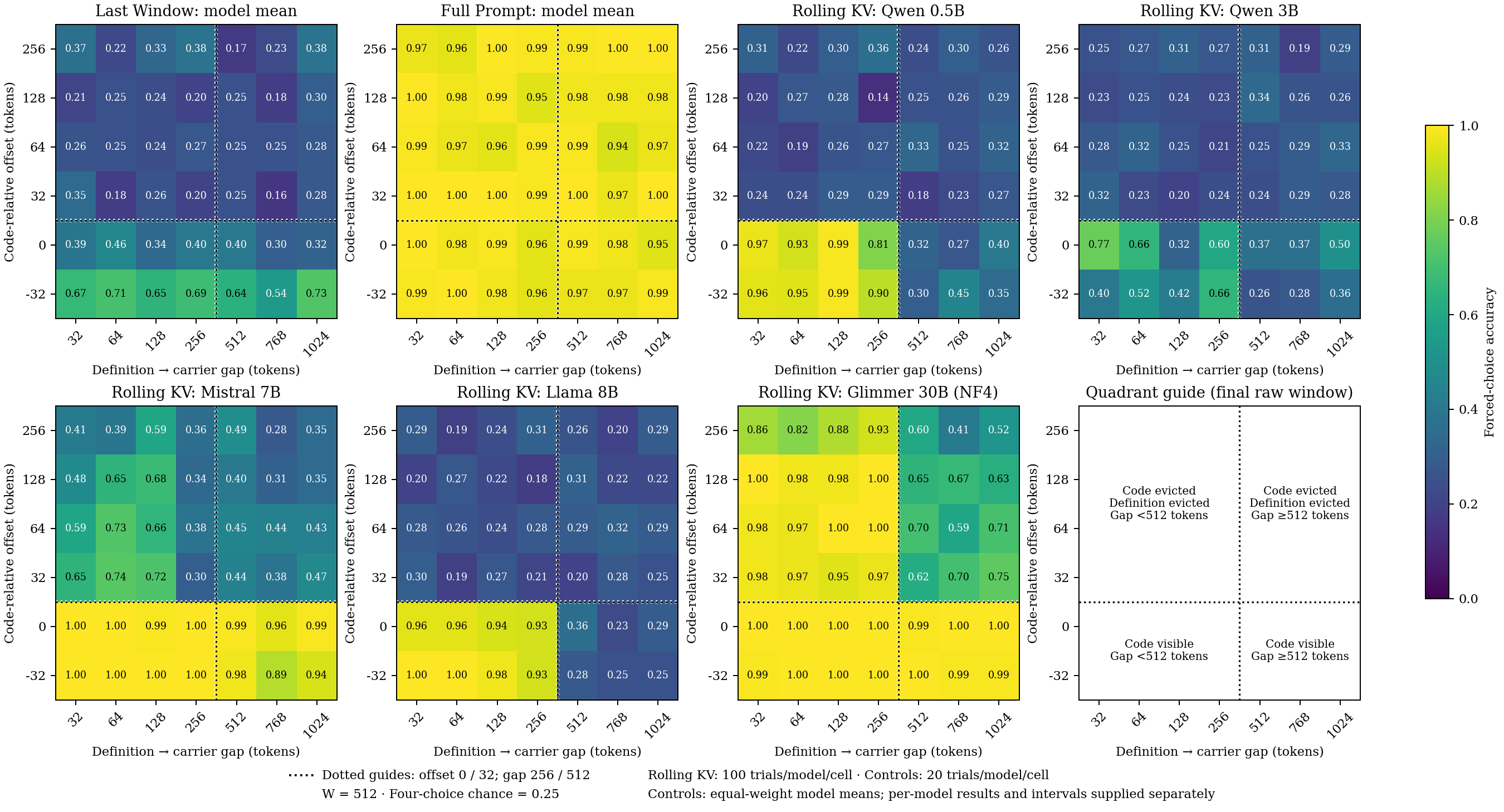}
    \caption{Experiment 3 forced-choice accuracy over the
    definition-to-carrier filler gap (horizontal axis) and the
    code-relative boundary offset (vertical axis), with $W=512$.
    The top two panels show equal-weight means of the five
    model-specific controls; the remaining panels show
    Rolling-KV accuracy for each model. Each Rolling-KV cell
    contains 100 trials per model, and each control cell
    contains 20 trials per model. Dotted lines separate
    offsets $0$ and $32$, and gaps $256$ and $512$.
    Positive offsets shown here place the code outside the
    final raw window and retained cache. Four-choice chance
    accuracy is $25\%$.}
    \label{fig:definition-carrier-grid}
\end{figure}

For each model and grid cell, Rolling-KV is evaluated on 100
trials. The Last Window and Full Prompt controls each use 20
prompts sampled from those saved Rolling-KV trials per model
and cell. Figure~\ref{fig:definition-carrier-grid} displays
the two controls as equal-weight means across models and the
Rolling-KV results separately for each model.

\paragraph{Results}
Full Prompt accuracy remains high across the grid, indicating
that the task is generally solvable when the definition and code
are available together. Once the code is absent from the final
raw window, Last Window accuracy is generally near the
four-choice chance level of $25\%$.

The dotted lines in Figure~\ref{fig:definition-carrier-grid}
divide each heat map into four regions. The horizontal line
falls between offsets $0$ and $32$: below it, the code is
visible in the final raw window; above it, the code has been
evicted. The vertical line falls between gaps of $256$ and
$512$ filler tokens, separating gaps shorter than $W$ from
gaps of at least $W$.

In the bottom-left region, the code is visible at the final
step and the definition-to-carrier filler gap is shorter than
one window. Four of the five models perform strongly under
Rolling-KV, although Qwen 3B is notably weaker than Qwen 0.5B.
In the bottom-right region, the code is still visible but the
gap is at least one window long. Mistral 7B and Glimmer 30B
remain highly accurate, while the other three models decline.

The upper regions test the stronger case in which the code
itself has been evicted. In the top-left region, the filler
gap is shorter than one window: Glimmer averages $95.4\%$
across these cells, while Mistral averages $54.2\%$. The
top-right region combines an evicted code with a gap of at
least $512$ filler tokens. It is the hardest setting, yet
Glimmer averages $62.9\%$ and Mistral $39.9\%$; the other
three models remain close to chance. These regional means
summarize the displayed cells, rather than a separately
sampled test.

\FloatBarrier

\section{Discussion}

The experiments address the central question of this study: whether
information remains usable after its source leaves a rolling KV window,
and for how long. They do not characterize all information transmitted
through the cache, but they reveal three settings in which Rolling-KV
retrieves a secret code that the Last Window control cannot reliably
identify. Experiments 1 and 2 probe relatively short distances around
the window boundary. Experiment 3 tests whether useful information can
persist across a definition-to-carrier gap extending beyond the retained
window size, as the dependency analysis in
Section~\ref{sec:information-propagation} suggests is possible.

\paragraph{Contextualized Representations as Carriers}
The first mode occurs when the secret code remains visible near the
end of the prompt but its defining instruction, such as ``The secret
code is:'', does not. Rolling-KV processed the code while that
instruction was available, so the retained KV states associated with
the code were computed in its context. Last Window sees the code
without that earlier definition and cannot reliably identify it as
the requested answer. We call this mode \emph{contextualized
representations as carriers}. All five models benefit from this mode
in the tested conditions.

\paragraph{Latent Information Relay}
The second mode occurs when both the defining instruction and the code
itself have left the final window. The Last Window input then contains
neither the answer nor the information identifying it as the secret.
Some models nevertheless recover the code with Rolling-KV. In
Experiments 1 and 2, Llama 3.1 shows substantial retrieval immediately
after the code's KV state is no longer directly accessible, but its
accuracy declines at larger offsets. Mistral 7B and Muse Glimmer
sustain high accuracy across the measured post-eviction offsets,
whereas the Qwen2.5 models decline much more sharply.

The successful Rolling-KV retrieval shows that information from the
earlier prompt remains usable through retained states after the code's
own KV entries have been evicted. We call this mode \emph{latent
information relay}. It is stronger than retrieval from a still-cached
code representation because a later retained state must support the
answer. This suggests a route toward recurrent-style memory during
streaming inference, although these experiments do not identify the
specific states or computations carrying the information.

\paragraph{Long-Range Transmission}
Experiment 3 extends this question by placing filler between the
definition and the later marker--code carrier. When this gap reaches
or exceeds the $512$-token window, the defining instruction is no
longer directly available as the carrier is processed. Mistral and
Glimmer remain highly accurate in this setting while the code is
still retained. Glimmer also retains above-chance retrieval in the
hardest region, where the gap is at least $512$ tokens and the code
itself has subsequently been evicted. These results show that useful
information can cross a window-scale gap and remain available beyond
the code's own eviction. Performance varies substantially by model
and by distance, so the existence of this path does not imply
reliable long-term memory in general.

\paragraph{Limitations}
Our evidence is behavioral: a model produces the correct answer
despite the source tokens being unavailable for direct attention at
the final step. It does not show how much of the original information
survives or whether the evicted text could be reconstructed. We
therefore distinguish task-relevant preservation from lossless memory.
The retained states need only carry consequences of the earlier
context sufficient for this retrieval task.

Failure to retrieve the code also does not prove that no information
about it remains. A retained state might contain information that
the model cannot use to answer the particular question. Conversely,
successful retrieval provides evidence that some surviving
information is functionally accessible, without identifying its
location or representation in the cache. Direct interventions on
retained states would be needed to examine that mechanism more
closely.

The experiments use a synthetic four-choice task, a fixed
$W=512$ cache, and a small set of models. Accuracy on this task does
not establish how much information a rolling cache can preserve in
natural dialogue, nor how long it remains useful as more content is
ingested. The models differ in size, architecture, training, and
precision, so their relative performance cannot isolate the cause
of their different behavior. Finally, the experiments measure
retrieval rather than directly tracing information through
individual layers or cached states. 

\paragraph{Future Work}
These experiments show that usable information can survive beyond a
fixed rolling KV window, but they do not show that a model will
reliably use this path. Mistral 7B and Muse Glimmer, the two models
with the strongest retrieval after code eviction, include local
sliding attention in their published architectures. The Qwen2.5
models tested here disable sliding attention, while Llama 3.1 uses
a standard attention architecture. This association is suggestive,
but the models differ in many other respects, and their published
descriptions do not establish that any was trained specifically to
preserve information after KV eviction.

A more direct test would train matched models with the same
architecture and data, varying whether training includes a rolling
cache and tasks whose relevant information must cross one or more
window boundaries. Evaluation on new markers, content, and distances
could then test whether such training improves generalizable
long-term retrieval. Interventions on the retained KV states could
also help identify where task-relevant information travels and when
it becomes inaccessible.

\paragraph{Safety}
A safety motivation for this work is understanding what information a
model can still use after its source has left the visible context.
Directly questioning a model can provide behavioral evidence about
information available from its internal processing. Fornasiere et
al.\ found that language models could identify where their activations
had been perturbed and, in some settings, distinguish the type of
perturbation \parencite{fornasiere2026activations}. Our experiments
similarly ask models to retrieve information encountered earlier during
inference. Successful retrieval shows that the information remains
usable for this task, though it does not reveal where or how it is
represented.

Identifying the path that carries this information requires a different
kind of investigation. Anthropic's circuit-tracing work maps features
and their interactions during particular model computations
\parencite{ameisen2025circuit}. Applying such methods to rolling
inference could help trace how information from an evicted source token
influences later retained states and, ultimately, the answer. This
would connect the behavioral effects observed here to a more specific
internal mechanism.

\section*{Author Contributions}
Tim led the research, designed and implemented the
experiments, analyzed the results, and drafted the manuscript.
Seth contributed to the development and discussion of
the ideas, wrote substantial portions of the several sections, ran experiments, 
and revised the manuscript. Both authors reviewed and approved
the final version.

\section*{AI-Assisted Tools}
The research questions, experimental design, results, and scientific
interpretation were developed by the authors. ChatGPT was used to assist
with manuscript editing, proofreading, and LaTeX formatting, and Codex
was used for coding assistance. The authors reviewed the resulting
manuscript and code and take responsibility for their content.

\section*{Code and Data Availability}
Code, experiment configurations, raw trial results, and figures are
available at
\url{https://github.com/jefferythewind/Thinking-Outside-the-Box}.







%
%
%





\printbibliography

@misc{arditi2024refusallanguagemodelsmediated,
      title={Refusal in Language Models Is Mediated by a Single Direction}, 
      author={Andy Arditi and Oscar Obeso and Aaquib Syed and Daniel Paleka and Nina Panickssery and Wes Gurnee and Neel Nanda},
      year={2024},
      eprint={2406.11717},
      archivePrefix={arXiv},
      primaryClass={cs.LG},
      url={https://arxiv.org/abs/2406.11717}, 
}

@article{jiang2023mistral,
  title   = {Mistral 7B},
  author  = {Jiang, Albert Q. and Sablayrolles, Alexandre and Mensch, Arthur
             and Bamford, Chris and Chaplot, Devendra Singh and de las Casas, Diego
             and Bressand, Florian and Lengyel, Gianna and Lample, Guillaume
             and Saulnier, Lucile and Lavaud, L{\'e}lio Renard
             and Lachaux, Marie-Anne and Stock, Pierre and Le Scao, Teven
             and Lavril, Thibaut and Wang, Thomas and Lacroix, Timoth{\'e}e
             and El Sayed, William},
  journal = {arXiv preprint arXiv:2310.06825},
  year    = {2023}
}

@misc{meta2026museglimmer,
  author       = {{Meta AI}},
  title        = {Introducing Muse Glimmer: An Open Agentic Model That Runs on Your Device},
  year         = {2026},
  month        = aug,
  howpublished = {\url{https://research.meta.ai/blog/introducing-muse-glimmer-open-agentic-model/}},
  note         = {Accessed: 2026-09-09}
}

@inproceedings{vaswani2017attention,
  title     = {Attention Is All You Need},
  author    = {Vaswani, Ashish and Shazeer, Noam and Parmar, Niki
               and Uszkoreit, Jakob and Jones, Llion and Gomez, Aidan N.
               and Kaiser, Lukasz and Polosukhin, Illia},
  booktitle = {Advances in Neural Information Processing Systems},
  volume    = {30},
  year      = {2017}
}

@article{gu2023mamba,
  title   = {Mamba: Linear-Time Sequence Modeling with Selective State Spaces},
  author  = {Gu, Albert and Dao, Tri},
  journal = {arXiv preprint arXiv:2312.00752},
  year    = {2023}
}

@article{elhage2021mathematical,
  title   = {A Mathematical Framework for Transformer Circuits},
  author  = {Elhage, Nelson and Nanda, Neel and Olsson, Catherine and
             Henighan, Tom and Joseph, Nicholas and Mann, Ben and
             Askell, Amanda and Bai, Yuntao and Chen, Anna and
             Conerly, Tom and DasSarma, Nova and Drain, Dawn and
             Ganguli, Deep and Hatfield-Dodds, Zac and Hernandez, Danny and
             Jones, Andy and Kernion, Jackson and Lovitt, Liane and
             Ndousse, Kamal and Amodei, Dario and Brown, Tom and
             Clark, Jack and Kaplan, Jared and McCandlish, Sam and
             Olah, Chris},
  journal = {Transformer Circuits Thread},
  year    = {2021},
  url     = {https://transformer-circuits.pub/2021/framework/index.html}
}

@article{meng2022locating,
  title   = {Locating and Editing Factual Associations in {GPT}},
  author  = {Meng, Kevin and Bau, David and Andonian, Alex and Belinkov, Yonatan},
  journal = {arXiv preprint arXiv:2202.05262},
  year    = {2022}
}

@article{xiao2023streamingllm,
  title   = {Efficient Streaming Language Models with Attention Sinks},
  author  = {Xiao, Guangxuan and Tian, Yuandong and Chen, Beidi and
             Han, Song and Lewis, Mike},
  journal = {arXiv preprint arXiv:2309.17453},
  year    = {2023}
}

@article{zhang2023h2o,
  title   = {H$_2$O: Heavy-Hitter Oracle for Efficient Generative Inference
             of Large Language Models},
  author  = {Zhang, Zhenyu and Sheng, Ying and Zhou, Tianyi and
             Chen, Tianlong and Zheng, Lianmin and Cai, Ruisi and
             Song, Zhao and Tian, Yuandong and R{\'e}, Christopher and
             Barrett, Clark and Wang, Zhangyang and Chen, Beidi},
  journal = {arXiv preprint arXiv:2306.14048},
  year    = {2023}
}

@article{dai2019transformerxl,
  title   = {Transformer-XL: Attentive Language Models Beyond a
             Fixed-Length Context},
  author  = {Dai, Zihang and Yang, Zhilin and Yang, Yiming and
             Carbonell, Jaime and Le, Quoc V. and Salakhutdinov, Ruslan},
  journal = {arXiv preprint arXiv:1901.02860},
  year    = {2019}
}

@article{rae2019compressive,
  title   = {Compressive Transformers for Long-Range Sequence Modelling},
  author  = {Rae, Jack W. and Potapenko, Anna and Jayakumar, Siddhant M.
             and Lillicrap, Timothy P.},
  journal = {arXiv preprint arXiv:1911.05507},
  year    = {2019}
}

@article{nvidia2025nemotronh,
  title   = {Nemotron-H: A Family of Accurate and Efficient Hybrid
             Mamba-Transformer Models},
  author  = {{NVIDIA}},
  journal = {arXiv preprint arXiv:2504.03624},
  year    = {2025}
}

@misc{qwen2025qwen25,
      title={Qwen2.5 Technical Report},
      author={Qwen Team},
      year={2025},
      eprint={2412.15115},
      archivePrefix={arXiv},
      primaryClass={cs.CL},
      url={https://arxiv.org/abs/2412.15115},
}

@article{dubey2024llama,
      title={The Llama 3 Herd of Models},
      author={Dubey, Abhimanyu and Jauhri, Abhinav and Pandey, Abhinav
              and Kadian, Abhishek and Al-Dahle, Ahmad and Letman, Aiesha
              and Mathur, Akhil and Schelten, Alan and Yang, Amy and Fan, Angela
              and others},
      year={2024},
      journal={arXiv preprint arXiv:2407.21783},
      url={https://arxiv.org/abs/2407.21783}
}

@article{ameisen2025circuit,
  author  = {Emmanuel Ameisen and Jack Lindsey and Adam Pearce and others},
  title   = {Circuit Tracing: Revealing Computational Graphs in Language Models},
  journal = {Transformer Circuits Thread},
  year    = {2025},
  url     = {https://transformer-circuits.pub/2025/attribution-graphs/methods.html}
}

@misc{fornasiere2026activations,
  title         = {Language Models Recognize Dropout and Gaussian Noise Applied to Their Activations},
  author        = {Fornasiere, Damiano and Bronzi, Mirko and Kitts, Spencer and Palmas, Alessandro and Bengio, Yoshua and Richardson, Oliver},
  year          = {2026},
  eprint        = {2604.17465},
  archivePrefix = {arXiv},
  primaryClass  = {cs.AI},
  url           = {https://arxiv.org/abs/2604.17465}
}

\end{document}